\documentclass[10pt,twocolumn,letterpaper]{article}

\usepackage{iccv}
\usepackage{times}
\usepackage{epsfig}
\usepackage{graphicx}
\usepackage{amsmath}
\usepackage{amssymb}
\usepackage{boldline}
\usepackage{multirow}
\usepackage{pifont}% http://ctan.org/pkg/pifont
\newcommand{\cmark}{\ding{51}}%
\newcommand{\xmark}{\ding{55}}%
\usepackage{subfig}

\usepackage[breaklinks=true,bookmarks=false]{hyperref}

\iccvfinalcopy % *** Uncomment this line for the final submission

\ificcvfinal\fi

\begin{document}

%%%%%%%%% TITLE
\title{Group-based Bi-Directional Recurrent Wavelet Neural Networks\\for Video Super-Resolution}

\author{Young-Ju Choi\\
Department of IT Engineering\\
Sookmyung Women's University, Seoul, Korea\\
{\tt\small yj.choi@ivpl.sookmyung.ac.kr}
\and
Young-Woon Lee\\
Department of Computer Engineering\\
Sunmoon University, Asan, Korea\\
{\tt\small yw.lee@ivpl.sookmyung.ac.kr}
\and
Byung-Gyu Kim\\
Department of IT Engineering\\
Sookmyung Women's University, Seoul, Korea\\
{\tt\small bg.kim@sookmyung.ac.kr}
}

\maketitle
\ificcvfinal\thispagestyle{empty}\fi

%%%%%%%%% ABSTRACT
\begin{abstract}
Video super-resolution (VSR) aims to estimate a high-resolution (HR) frame from a low-resolution (LR) frames.
The key challenge for VSR lies in the effective exploitation of spatial correlation in an intra-frame and temporal dependency between consecutive frames.
However, most of the previous methods treat different types of the spatial features identically and extract spatial and temporal features from the separated modules.
It leads to lack of obtaining meaningful information and enhancing the fine details.
In VSR, there are three types of temporal modeling frameworks: 2D convolutional neural networks (CNN), 3D CNN, and recurrent neural networks (RNN).
Among them, the RNN-based approach is suitable for sequential data.
Thus the SR performance can be greatly improved by using the hidden states of adjacent frames.
However, at each of time step in a recurrent structure, the RNN-based previous works utilize the neighboring features restrictively.
Since the range of accessible motion per time step is narrow, there are still limitations to restore the missing details for dynamic or large motion.
In this paper, we propose a group-based bi-directional recurrent wavelet neural networks (GBR-WNN) to exploit the sequential data and spatio-temporal information effectively for VSR.
The proposed group-based bi-directional RNN (GBR) temporal modeling framework is built on the well-structured process with the group of pictures (GOP).
We propose a temporal wavelet attention (TWA) module, in which attention is adopted for both spatial and temporal features.
Experimental results demonstrate that the proposed method achieves superior performance compared with state-of-the-art methods in both of quantitative and qualitative evaluations.
\end{abstract}

%%%%%%%%% BODY TEXT
\section{Introduction}

Super-resolution (SR) is a traditional problem in low-level vision field.
The goal of SR is to reconstruct a high-resolution (HR) image from the corresponding low-resolution (LR) image.
Therefore, finding missing edge and texture details in LR image plays an important role in the SR.
The SR imaging technique is widely used in various computer vision applications such as medical, satellite, surveillance, and low-bitrate media imaging systems.
Moreover, with the growth of display industries, the SR has become more crucial in recent years.

The SR problem can be separated into single-image super-resolution (SISR)~\cite{Dong2014Learning,kim2016Accurate,Ledig2017Photo,Lim2017Enhanced,Shi2016Real,Zhang2018Residual}, multi-image super-resolution (MISR)~\cite{faramarzi2013unified,garcia2012super,liu2013bayesian}, and video super-resolution (VSR)~\cite{Caballero2017real,choi2020wavelet,haris2019recurrent,kappeler2016video,Sajjadi2018frame,tian2020tdan,wang2019edvr}.
Given a LR video consisting of ${(2N+1)}$ LR frames ${\{LR_{t-N}, ..., LR_{t}, ..., LR_{t+N}\}}$, the target frame is ${LR_{t}}$.
The SISR resolves each of the video frames independently.
However, this technique do not consider the temporal information from the other frames, which is very inefficient.
The MISR utilizes the temporal details from the neighboring frames and fuses them for super-resolving ${LR_{t}}$.
In MISR, however, the frames are aligned without any technique related to temporal smoothness.
This inaccurate motion alignment cause a result containing some discontinuous regions.
In contrast with the SISR and MISR, the VSR exploits temporal dependency among frames while maintaining a sequential characteristic.

With the success of the convolutional neural networks (CNN) in computer vision tasks such as image classification~\cite{Krizhevsky2012imagenet} and object detection~\cite{girchick2014rich}, CNN has also been successfully applied to VSR task.
By learning the non-linear LR-to-HR mapping function directly, the performance of reconstruction accuracy and visual quality was remarkably enhanced.
Earlier deep learning-based VSR methods~\cite{dai2015dictionary,liao2015video,takeda2009super} have designed as a simple extension of the SISR.
It means that the temporal information among video frames is not considered properly.

To exploit more temporal dependency between consecutive frames, some studies \cite{Caballero2017real,kappeler2016video,Sajjadi2018frame} have used an optical flow-based explicit motion estimation and compensation process.
% These works have devised a temporal alignment module based on optical flow estimation to warp neighboring frames to the target frame. 
However, it is difficult to estimate accurate motion in the case of a sequence including occlusions and large motions.
% An imprecise motion estimation and compensation would result in distortions and super-resolution performance degradation.
To address the aforementioned problem, recent methods \cite{jo2018deep,tian2020tdan} used implicit motion compensation. 
The latest state-of-the-art methods \cite{choi2020wavelet,Fan2019empirical,Purohit2019Mixed,wang2019edvr} have designed architectures with more elaborated pipelines and multiple stages to extract spatio-temporal feature and reconstructed target frame.

Furthermore, the VSR methods can be divided into three categories based on a temporal modeling framework: 2D CNN~\cite{Caballero2017real,choi2020wavelet,Purohit2019Mixed,tian2020tdan,wang2020deep,wang2019edvr}, 3D CNN~\cite{Fan2019empirical,jo2018deep}, and recurrent neural networks (RNN)~\cite{haris2019recurrent,huang2017video,Sajjadi2018frame}.
Most of the recent VSR approaches have used a frame concatenation and several stacked 2D CNN layers to extract spatial and temporal information.
In this case, although an additional temporal feature extraction module is constructed, there is still limitation to represent dynamic motion because input frames are concatenated together.
In VSR, input frames are stacked along the temporal axis.
To alleviate the problem with 2D CNN, some methods utilized 3D CNN to apply to the stacked frames directly.

The 3D CNN could extract both spatial and temporal information within a temporal sliding window simultaneously.
However, 3D CNN has higher computational complexity and needs a much larger memory than 2D CNN.
The RNN is effective in dealing with sequential data, so it could be employed for VSR.
In RNN structure, the frames is fed into convolution layers in temporal order.
With a RNN-based approach, a performance can be greatly improved because it is possible to use the hidden states of adjacent frames.

Exploiting both intra-frame spatial correlations and inter-frame temporal dependencies between consecutive frames plays an important role in the VSR.
From temporal point of view, the previous uni-directional RNN-based temporal modeling frameworks~\cite{haris2019recurrent,Sajjadi2018frame} have considered only the previous hidden states.
In the previous bi-directional RNN-based temporal modeling framework~\cite{huang2017video}, the previous and future hidden states at ${t-1}$ and ${t+1}$ time step have connected to the hidden state at current time step.
Because they dealt with the narrow range of motion representation, they may fail to address large and dynamic motions on the sequence.

% All of the independent frames in a sequence contain low and high-frequency components.
In an image, the low-frequency component describes the basic background information and the high-frequency component represents the edge and texture details.
From spatial point of view, most previous methods have handled the features identically or simply combined the edge map.
Therefore the HR output image is lack meaningful information.
Furthermore, since most previous model architectures have considered spatial and temporal feature extractor separately, there is a discontinuity between the two extracted features.
For better feature extraction, it is preferable that the spatial and temporal information is extracted together on single module.
Recently, Wang \etal~\cite{wang2019edvr} have proposed the temporal and spatial attention (TSA) module to extract a spatio-temporal feature.
The TSA module sequentially has performed the temporal attention and the spatial attention with several convolution layers.
However, the TSA module may have limitation because they have treated the low-frequency and high-frequency features equally.

In this paper, we propose a group-based bi-directional recurrent wavelet neural networks (GBR-WNN) to exploit the sequential data and spatio-temporal dependencies effectively for VSR.
The cores of GBR-WNN are (1) a temporal modeling framework, called group-based bi-directional RNN (GBR), and (2) a temporal wavelet attention (TWA) module.

The proposed GBR framework can cover a wide range of motion by utilizing the previous and future hidden states of time step ${\{t-N, ..., t, ..., t+N\}}$ within a defined group of pictures (GOP).
The proposed TWA module can extract an elaborate spatio-temporal feature by generating two attention maps for spatial and temporal information, respectively.
In the proposed TWA module, we apply discrete wavelet transform (DWT) for spatial attention to utilize the advantage of decomposing a feature to multiple features with frequency properties.
The scheme of temporally weighted features between frames can reduce discontinuities along the time axis.
Also, the weighted feature through four DWT attention maps with different frequency characteristics can keep the inherent precise and sharp attributes.

This paper is organized as follows: In section 2, we introduce the related works. In section 3, we present the proposed methodology. The experimental results are shown in section 4. Finally, section 5 will make a conclusion for this paper.

%-------------------------------------------------------------------------

\begin{figure*}
\begin{center}
\includegraphics[width=0.99\linewidth]{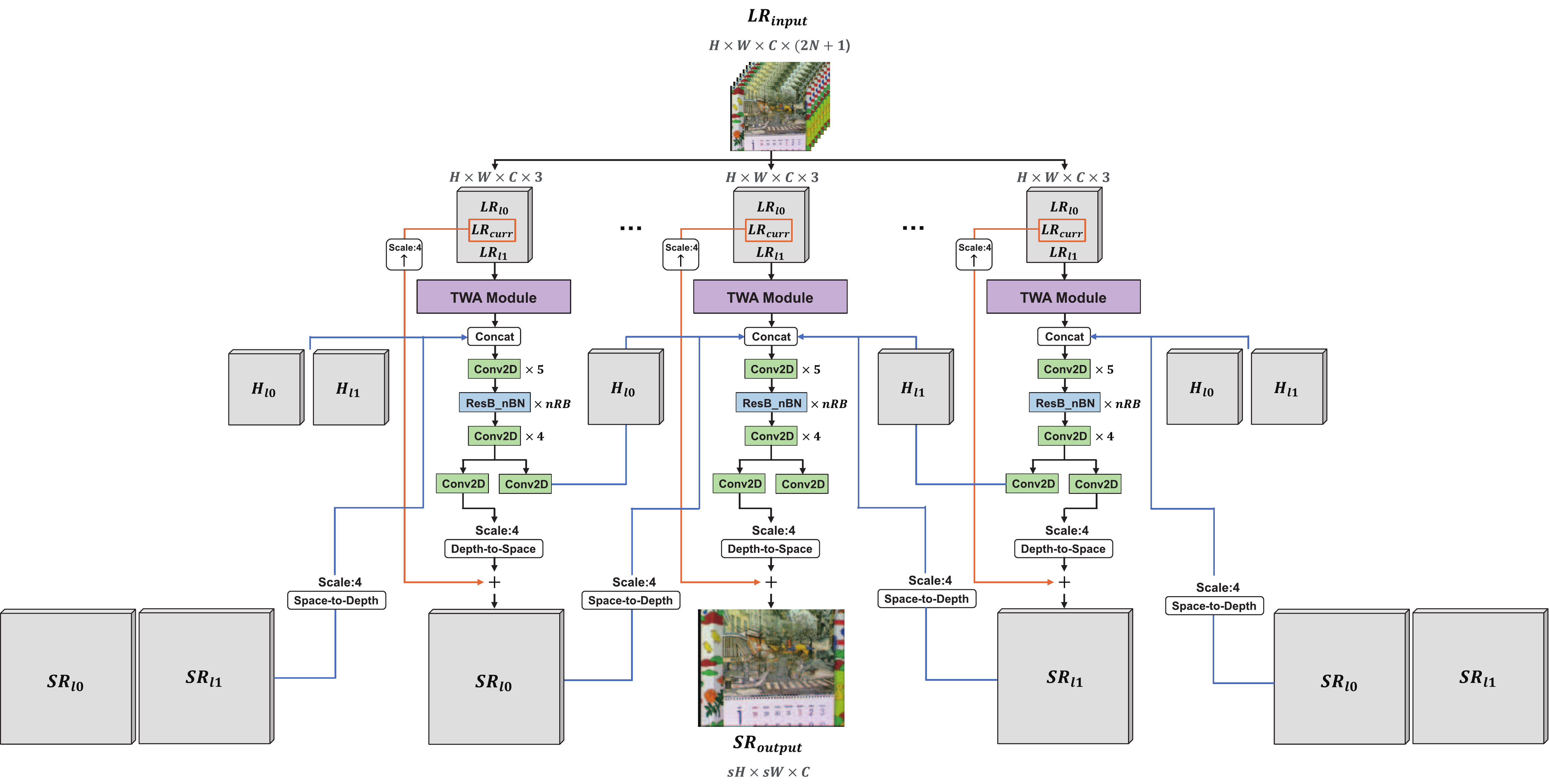}
\end{center}
   \caption{An overview of the proposed GBR-WNN.}
\label{fig1}
\end{figure*}

%------------------------------------------------------------------------
\section{Related Work}

%-------------------------------------------------------------------------
\subsection{Single Image Super-Resolution}

As an earliest work of deep learning-based SISR, Dong \etal~\cite{Dong2014Learning} proposed a super-resolution convolutional neural network (SRCNN). 
The SRCNN is a relatively shallow network.
Kim \etal~\cite{kim2016Accurate} later developed a deeper network with residual learning called very deep super-resolution (VDSR).
After that, an efficient sub-pixel convolutional neural network (ESPCN) proposed by Shi \etal~\cite{Shi2016Real} to reduce computational complexity with keeping a deeper network.
Also, Ledig \etal~\cite{Ledig2017Photo} proposed a super-resolution using a generative adversarial network (SRGAN), which was a model focusing to high frequency details.
Based on~\cite{Ledig2017Photo}, Lim \etal~\cite{Lim2017Enhanced} proposed an enhanced deep super-resolution network (EDSR), which modified the residual module by removing batch normalization.
Recently, much deeper CNN, including residual dense network (RDN)~\cite{Zhang2018Residual}, DBPN~\cite{haris2018deep}, RCAN~\cite{zhang2018image} were then introduced.
They outperformed previous networks by a substantial margin.

%-------------------------------------------------------------------------
\subsection{Video Super-Resolution}

%-------------------------------------------------------------------------
\subsubsection{2D CNN}

By inspiring of SRCNN in SISR, some researches in deep learning based VSR has begun.
Kappelar \etal~\cite{kappeler2016video} proposed a two-step framework consisting of flow estimation and frame warping, namely, video super-resolution with convolutional neural networks (VSRNet).
Caballero \etal~\cite{Caballero2017real} introduced the first end-to-end CNN for VSR, called as video super-resolution using an efficient sub-pixel convolutional neural network (VESPCN), which has been trained flow estimation and spatio-temporal networks.
The aforementioned methods used the optical flow to estimate the motions between frames and perform warping.
However, a motion estimation by utilizing the optical flow mechanism could be inaccurate in the case of a sequence including occlusion and large motion.
To address the issue, most of the latest models~\cite{choi2020wavelet,Purohit2019Mixed,tian2020tdan,wang2019edvr} handle the problem by using implicit motion compensation and surpass the optical flow-based methods.
Some of the latest methods~\cite{li2020learning,wang2020deep} still estimated optical flow and utilized the warped feature in a sophisticated multi-stage CNN architecture.

%-------------------------------------------------------------------------
\subsubsection{3D CNN}

Typically, a 3D CNN is more appropriate to extract spatio-temporal features than 2D CNN in a sequence.
Jo \etal~\cite{jo2018deep} proposed a VSR network estimating dynamic upsampling filters (DUF) with stacked 3D convolutional layers.
DUF handle the problem of explicit motion compensation by using implicit motion compensation and surpass the flow-based methods.
Fan \etal~\cite{Fan2019empirical} proposed 3D CNN-based architecture for deep video restoration, namely, wide-activated 3D convolutional network for video restoration (WDVR).

%-------------------------------------------------------------------------
\subsubsection{RNN}

RNN architecture is suitable for training a sequential data such as video. 
Sajjadi \etal~\cite{Sajjadi2018frame} proposed the frame-recurrent architecture called FRVSR to use previously inferred HR estimates for the SR of next frames.
After that, Haris \etal~\cite{haris2019recurrent} proposed the recurrent back-projection network (RBPN), which collected a temporal and spatial information from frames surrounding the target frame.
As a bi-directional RNN approach, Huang \etal~\cite{huang2017video} proposed a bi-directional recurrent convolutional network by using recurrent and 3D feedforward convolutions.

%-------------------------------------------------------------------------

%-------------------------------------------------------------------------
\section{Methodology}

%-------------------------------------------------------------------------
\subsection{Overview}

In this section, we introduce the overall system of our group-based bi-directional recurrent wavelet neural networks (GBR-WNN).
The proposed GBR-WNN consists of two core components: a temporal modeling framework, namely, group-based bi-directional RNN (GBR) and a feature extraction module which is called temporal wavelet attention (TWA).

The proposed network is illustrated in Figure~\ref{fig1}.
Given ${2N+1}$ consecutive LR input frames ${LR_{input}=\{LR_{t-N}, ..., LR_t, ..., LR_{t+N}\}}$, we denote the central LR frame ${LR_t}$ as the target frame and the other frames as neighboring frames with size of ${H \times W}$, where ${H}$ is height and ${W}$ is width.
The goal of VSR is to estimate a HR target frame ${SR_t}$, which is close to the ground truth frame ${HR_t}$ with size of ${\small{s}H \times \small{s}W}$, where ${s}$ is scaling factor.
% The overall framework of our network is built on a bi-directional RNN with group-based SR procedure.

In the GBR framework, the availability of temporal information is checked by exploiting the previous and future hidden states within the scope of a group of pictures (GOP).
The input LR frames are fed to the TWA module to extract elaborate spatio-temporal features based on 2D discrete wavelet transform (DWT)~\cite{Mallat1989theory}.
The output features after the TWA module pass through a reconstruction and upsampling module.
Our reconstruction module is designed with several 2D CNN layers and 2D residual blocks wherein no batch normalization units in~\cite{Lim2017Enhanced}.
The predicted HR residual frame is obtained by adopting the depth-to-space transformation in~\cite{Shi2016Real}.
Finally, the HR estimated frame ${SR_t}$ is obtained by adding the predicted residual frame to a direct upsampled LR frame ${LR_t}$.

%-------------------------------------------------------------------------
\subsection{Group-based Bi-directional RNN}

\begin{figure}
\begin{center}
\includegraphics[width=0.95\linewidth]{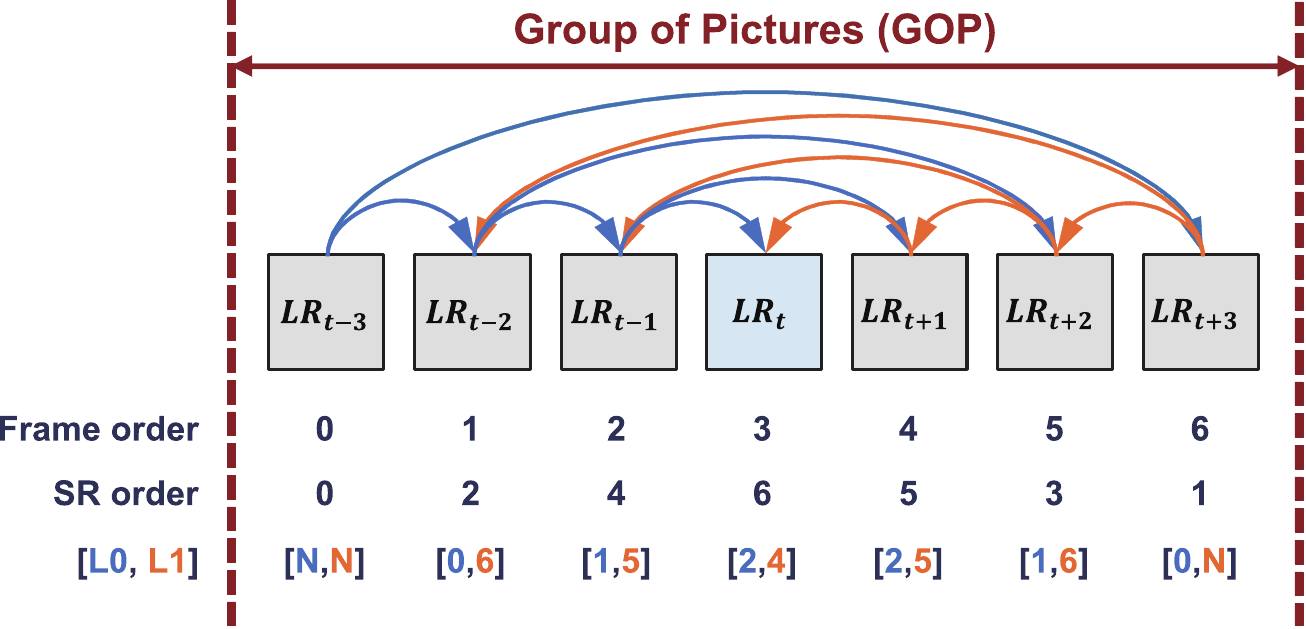}
\end{center}
   \caption{The structure of the proposed GBR framework.}
\label{fig2}
\end{figure}

In case of a sequence containing large motion, the far away neighboring frames are likely to have the missing details of the target frame.
Therefore, if we utilize a wide range of motion, we are able to have good chance to improve the performance in VSR.
In order to manage the hidden states of neighboring frames as well as the LR inputs of those dynamically, we propose the GBR temporal modeling framework.

The structure of the GPR framework is shown in Figure~\ref{fig2}.
We define the ${(2N+1)}$ LR input frames as the group of pictures (GOP).
In a GOP, each of the LR frames ${\{LR_{t-N}, ..., LR_t, ..., LR_{t+N}\}}$ becomes an current target LR frame ${LR_{curr}}$ in SR order.
For effective structure to estimate the central HR target frame ${SR_t}$, the SR order is gradually directed toward the center.
As the previous and future hidden states, and SR frames, we apply the features of the frames already predicted at the current point.
The time steps of the previous and future neighboring features are illustrated in Figure~\ref{fig2}, where ${L0}$, ${L1}$, and ${N}$ are previous picture list, future picture list, and none, respectively.
At the time step ${t-3}$ for the first target frame, the previous LR frame ${LR_{l0}}$ and the previous and future estimations ${H_{l0}}$, ${H_{l1}}$, ${SR_{l0}}$, and ${SR_{l1}}$ are initialized with zero. 
At the time step ${t+3}$ for the last target frame, the future LR frame ${LR_{l1}}$ and the future estimations ${H_{l1}}$ and ${SR_{l1}}$ are initialized with zero.

At every time step, the GBR-WNN uses following equations to estimate output SR frame at the current point.
The TWA module takes three LR frames ${LR_{l0}}$, ${LR_{curr}}$, and ${LR_{l1}}$ as input to produce the temporal wavelet feature ${TWF_{curr}}$:
\begin{equation}\label{eq1}
{TWF_{curr}=f_{twa}([LR_{l0},LR_{curr},LR_{l1}]),}
\end{equation}
where ${f_{twa}(\cdot)}$ and ${[\cdot,\cdot,\cdot]}$ denote the mapping function for TWA module and concatenation, respectively.
We adopt the space-to-depth~\cite{Shi2016Real} transformation to match the shape of the previous and future SR frames ${SR_{l0}}$ and ${SR_{l1}}$ to LR frames with scaling factor ${s}$:
\begin{equation}\label{eq2}
{\left\{\begin{array}{l}{SR_{l0}^{s \rightarrow d}=f_{s \rightarrow d}(SR_{l0})}, \\ 
{SR_{l1}^{s \rightarrow d}=f_{s \rightarrow d}(SR_{l1})}, \\ \end{array}\right.}
\end{equation}
where ${f_{s \rightarrow d}(\cdot)}$, ${SR_{l0}^{s \rightarrow d}}$, and ${SR_{l1}^{s \rightarrow d}}$ denote the space-to-depth mapping function, the reshaped previous and future SR frames, respectively.
Then, to further enhance the feature, concatenated feature with temporal wavelet feature, previous and future hidden states and SR frames are fed into reconstruction module.
This process can be represented as
\begin{equation}\label{eq3}
{H_{curr}=f_{rec}([TWF_{curr},H_{l0},H_{l1},SR_{l0}^{s \rightarrow d},SR_{l1}^{s \rightarrow d}]),}
\end{equation}
where ${f_{rec}(\cdot)}$ denotes the mapping function of reconstruction module.
${H_{curr}}$ represents the extracted current hidden state.
For the upsampling process, the depth-to-space mapping function ${f_{d \rightarrow s}(\cdot)}$ is applied to hidden state ${H_{curr}}$ with scaling factor ${s}$.
Lastly, output SR frame ${SR_{curr}}$ of each time step is obtained by global residual learning using the estimated HR residual frame and the upsampled LR target frame:
\begin{equation}\label{eq4}
{SR_{curr}=f_{d \rightarrow s}(H_{curr})+f_{up}^{\times s}(LR_{curr}),}
\end{equation}
where ${f_{up}^{\times s}(\cdot)}$ denotes the bilinear upsampling function with scaling factor ${s}$.
When the frame order of the current target frame is smaller than the frame order of next target frame, the current hidden state ${H_{curr}}$ becomes the previous hidden state ${H_{l0}}$ and the current SR frame ${SR_{curr}}$ becomes the previous SR frame ${SR_{l0}}$.
In contrast, if the frame order of the current target frame is bigger than the frame order of next target frame, the current hidden state ${H_{curr}}$ becomes the future hidden state ${H_{l1}}$ and the current SR frame ${SR_{curr}}$ becomes the future SR frame ${SR_{l1}}$.

%-------------------------------------------------------------------------
\subsection{Temporal Wavelet Attention}

\begin{figure}
\begin{center}
\includegraphics[width=0.6\linewidth]{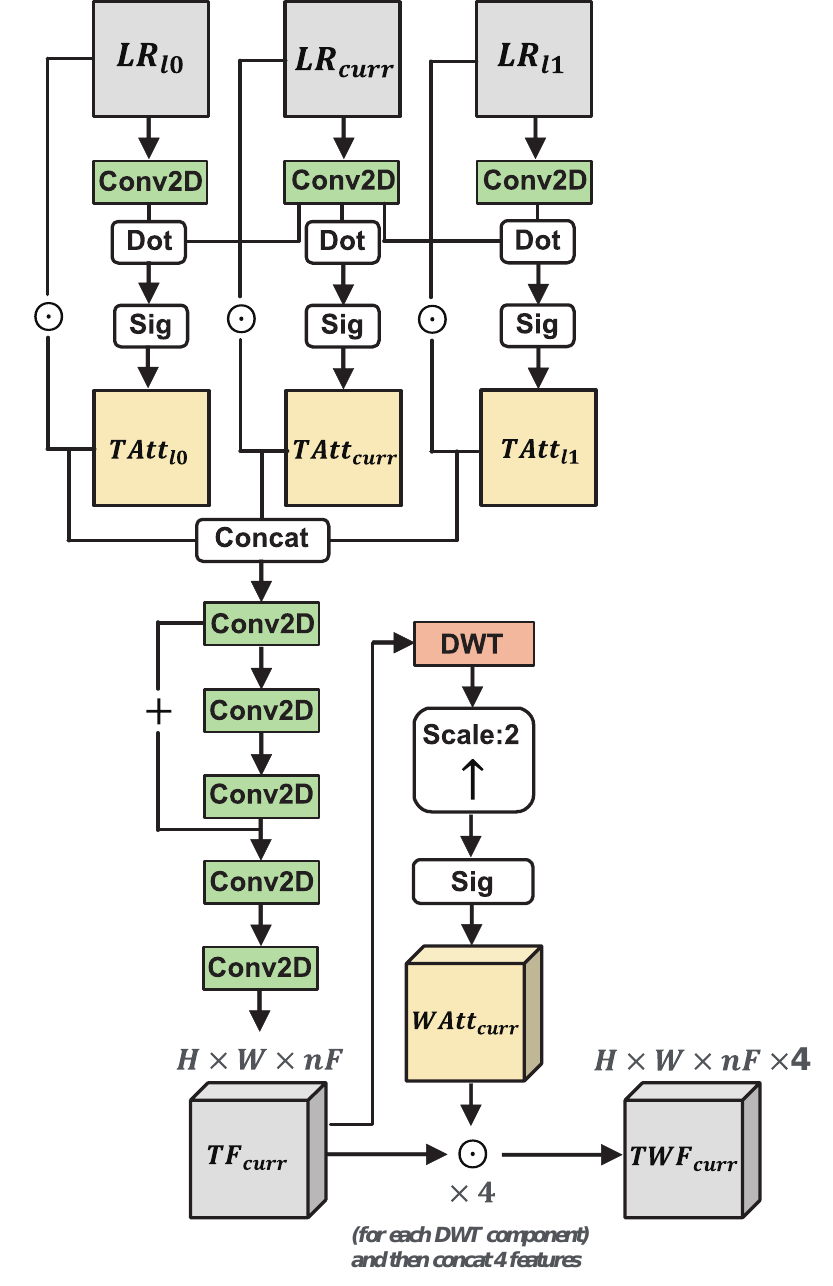}
\end{center}
   \caption{The network architecture of the proposed TWA module.}
\label{fig3}
\end{figure}

\begin{figure}
\begin{center}
\includegraphics[width=0.8\linewidth]{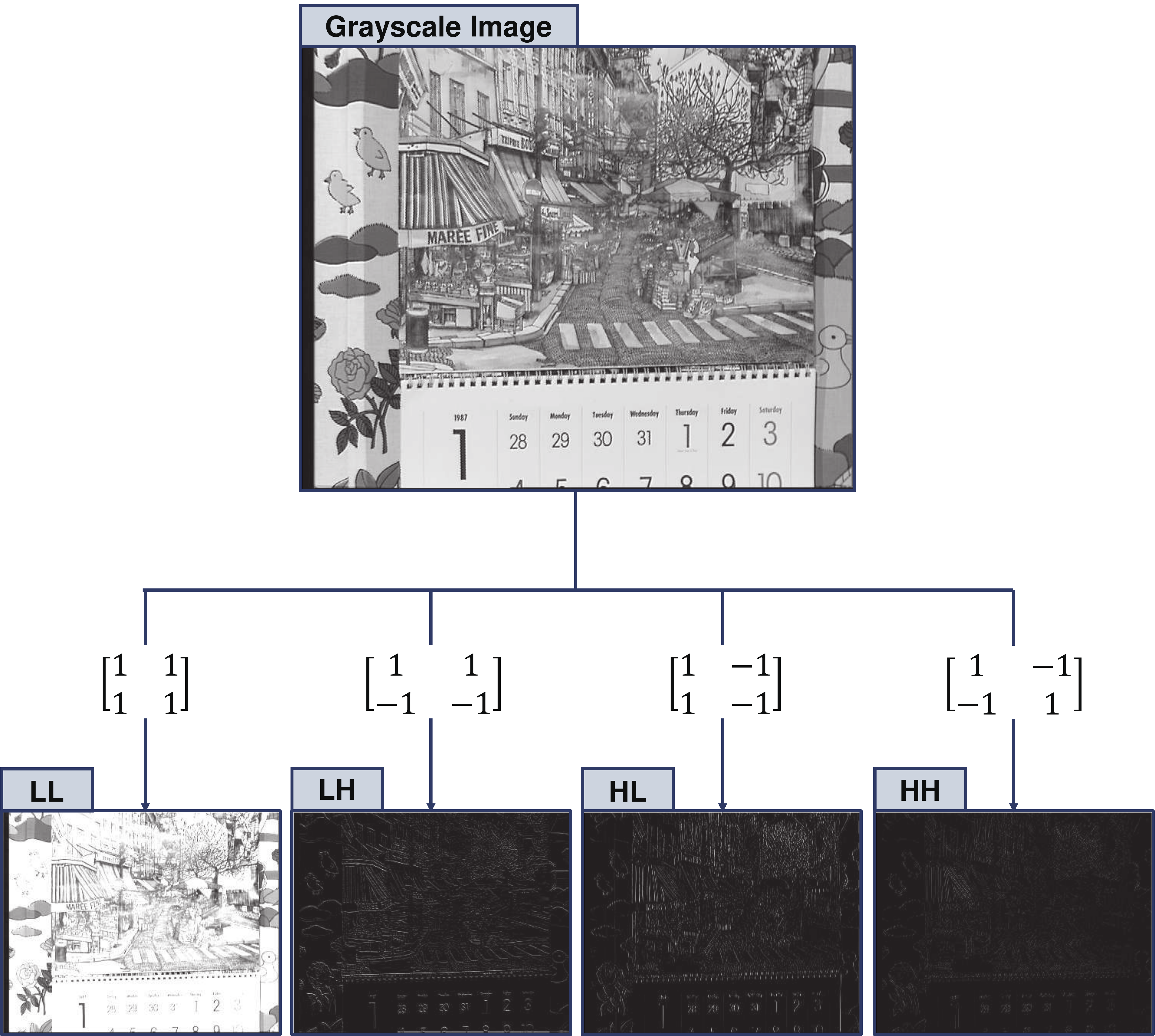}
\end{center}
   \caption{An example of the 2D Haar discrete wavelet transform (DWT).}
\label{fig4}
\end{figure}

Extracting fine feature of both intra-frame spatial and inter-frame temporal information is very important to improve the quality of output SR frame.
To obtain the spatio-temporal feature, we design the TWA module based on the 2D Haar DWT and attention mechanism, as illustrated in Figure~\ref{fig3}.
The key role of the TWA is generating the temporal attention map and wavelet attention map.
By utilizing temporal attention map, we can restore the missing feature from the neighboring frames with different degrees of motion information.
Meanwhile, we can strengthen the edge and texture details by applying the spatial attention map.

Each of the three LR frames are fed into the CNN embedding layer to increase the number of features:
\begin{equation}\label{eq5}
{\left\{\begin{array}{l}{E_{l0}=f_{emb}(LR_{l0})}, \\ 
{E_{curr}=f_{emb}(LR_{curr})}, \\ 
{E_{l1}=f_{emb}(LR_{l1})}, \\ \end{array}\right.}
\end{equation}
where ${f_{emb}(\cdot)}$ denotes the embedding function.
The embedded LR features of three LR frames are ${E_{l0}}$, ${E_{curr}}$, and ${E_{l1}}$, respectively.
The temporal attention features ${TAtt_{l0}}$, ${TAtt_{curr}}$, and ${TAtt_{l1}}$ for three LR frames can be calculated as
\begin{equation}\label{eq6}
{\left\{\begin{array}{l}{TAtt_{l0}=sigmoid(E_{l0} \boldsymbol{\cdot} E_{curr})}, \\ 
{TAtt_{curr}=sigmoid(E_{curr} \boldsymbol{\cdot} E_{curr})}, \\ 
{TAtt_{l1}=sigmoid(E_{l1} \boldsymbol{\cdot} E_{curr})}, \\ \end{array}\right.}
\end{equation}
where ${\boldsymbol{\cdot}}$ and ${sigmoid(\cdot)}$ denote the dot product and the sigmoid activation function. 
After that, the temporal attention maps are then multiplied to the original LR frames:
\begin{equation}\label{eq7}
{\left\{\begin{array}{l}{\tilde{E}_{l0}=LR_{l0} \odot TAtt_{l0}}, \\ 
{\tilde{E}_{curr}=LR_{curr} \odot TAtt_{curr}}, \\ 
{\tilde{E}_{l1}=LR_{l1} \odot TAtt_{l1}}, \\ \end{array}\right.}
\end{equation}
where ${\odot}$ denotes the element-wise multiplication.
The weighted temporal embedded features ${\tilde{E}_{l0}}$, ${\tilde{E}_{curr}}$, and ${\tilde{E}_{l1}}$ are then concatenated.  
Then, the current temporal feature ${TF_{curr}}$ can be represented as
\begin{equation}\label{eq8}
{TF_{curr}=f_{fusion}([\tilde{E}_{l0},\tilde{E}_{curr},\tilde{E}_{l1}]),}
\end{equation}
where ${f_{fusion}(\cdot)}$ denotes the fusion process function based on several CNN layers.

For extracting the spatial feature, we use the 2D Haar DWT.
An example of the 2D Haar DWT is shown in Figure~\ref{fig4}.
An image can be decomposed into four sub-band images using four 2D DWT filters.
The low-pass filter ${f_{LL}}$ means approximation of image and high-pass filters ${f_{HL}, \;f_{LH}, \;and \;f_{HH}}$ mean vertical, horizontal, and diagonal edge of image, respectively.
The DWT filters used in this paper are defined as
\begin{equation}\label{eq9}
\begin{array}{c@{\qquad}c}
f_{LL} = \begin{bmatrix} 1 & 1 \\ 1 & 1 \end{bmatrix},\;
f_{HL} = \begin{bmatrix} 1 & -1 \\ 1 & -1 \end{bmatrix},\\[10pt]
f_{LH} = \begin{bmatrix} 1 & 1 \\ -1 & -1 \end{bmatrix},\;
f_{HH} = \begin{bmatrix} 1 & -1 \\ -1 & 1 \end{bmatrix}.
\end{array}
\end{equation}

The estimated temporal feature ${TF_{curr}}$ can be decomposed into four sub-band components by using 2D DWT.
The wavelet attention map ${WAtt_{curr}}$ can be generated by the upsampling function and sigmoid activation function.
This process can be expressed as
\begin{equation}\label{eq10}
{WAtt_{curr}=sigmoid(f_{up}^{\times 2}(DWT(TF_{curr}))).}
\end{equation}
Because the wavelet attention map ${WAtt_{curr}}$ is composed of four components, the size of map is ${H \times W \times nF \times 4}$, where ${nF}$ means number of features.
Finally, by multiplying temporal feature ${TF_{curr}}$ and each DWT component of wavelet attention map ${[WAtt_{curr}^{LL},WAtt_{curr}^{HL},WAtt_{curr}^{LH},WAtt_{curr}^{HH}]}$, the temporal wavelet feature ${TWF_{curr}}$ can be achieved as
\begin{equation}\label{eq11}
{\left\{\begin{array}{l}{TWF_{curr}^{LL}=TF_{curr} \odot WAtt_{curr}^{LL}}, \\ 
{TWF_{curr}^{HL}=TF_{curr} \odot WAtt_{curr}^{HL}}, \\
{TWF_{curr}^{LH}=TF_{curr} \odot WAtt_{curr}^{LH}}, \\
{TWF_{curr}^{HH}=TF_{curr} \odot WAtt_{curr}^{HH}}, \\ \end{array}\right.}
\end{equation}
\begin{equation}\label{eq12}
{TWF_{curr}=[TWF_{curr}^{LL},TWF_{curr}^{HL},TWF_{curr}^{LH},TWF_{curr}^{HH}].}
\end{equation}

%-------------------------------------------------------------------------

%-------------------------------------------------------------------------
\section{Experimental Results}
%-------------------------------------------------------------------------
\subsection{Datasets and Implementation Details}
In this paper, we used the Vimeo-90K~\cite{Xue2017vimeo} dataset which is a large and diverse data set with high-quality frames and a range of motion types for training. 
The resolution of each sample in the Vimeo-90K dataset is ${448 \times 256}$.
For training, we used 3 channel patches of size ${64 \times 64}$ as inputs.
We augment the training data with random horizontal flips and ${90^{\circ}}$ rotations.
We evaluated our methods on the Vid4~\cite{liu2013vid4} dataset.
The Vid4 dataset consists of the four test sequences, walk, foliage, city, and calendar, commonly reported in recent methods.
% Therefore, we compared the proposed GBR-WNN with five state-of-the-art VSR models and Bicubic using Vid4 dataset.

In all our experiments, the scaling factor ${s}$ of SR was set to 4.
For evaluation, we use peak signal-to-noise ratio (PSNR) to test the quality of each frame.
The overall PSNR values of each video clip were then calculated by aggregating PSNRs over all frames in a video clip.
Finally, the overall PSNR values of whole Vid4 dataset were then calculated by averaging PSNRs of video clips.

For our GBR-WNN, the network takes seven consecutive frames ${(i.e.,\; N=3)}$ as inputs.
Also, the number of feature in each residual block was set to 128.
To train our network, we used Charbonnier penalty function~\cite{Lai2017Deep} for loss function:
\begin{equation}\label{loss}
{L=\sqrt{\left \| \hat{O_{t}}-O_{t} \right \|^{2}+\varepsilon^{2} },}
\end{equation}
where ${\varepsilon}$ set to ${1 \times 10^{-3}}$.
We used Adam optimizer~\cite{Kingma2015adam} and initially set learning rate to ${4 \times 10^{-4}}$.
The number of iterations was set to 600K.

The proposed GBR-WNN was implemented in PyTorch on a PC with 8 NVIDIA Tesla V100 16GB GPUs.
We trained with setting the size of global mini-batch to 128, which means that the size of mini-batch for each GPU was set to 16. 

%-------------------------------------------------------------------------

\begin{table*}[!t]
\caption{Quantitative comparison on Vid4 for 4${\times}$ video SR on Y channel. {\textcolor{red}{Red}} and {\textcolor{blue}{blue}} indicates the best and the second best performance, respectively.\label{table1}}
\begin{center}
\begin{tabular}{|c||c|c|c|c||c||c|}
\hline
\multirow{2}{*}{Method} & \multicolumn{5}{c|}{PSNR (dB)}                    & \multirow{2}{*}{Params.} \\ \cline{2-6}
                        & Calendar & City  & Foliage & Walk  & Average &                          \\ \hline\hline
Bicubic                 & 20.45    & 25.22 & 23.57   & 26.27 & 23.88   & -                        \\ \hline
SOF-VSR~\cite{wang2020deep}                 & 22.66    & 26.94 & 25.45   & 29.18 & 26.06   & 1.0M                     \\ \hline
WAEN~\cite{choi2020wavelet}                    & 23.81    & 27.61 & 26.00   & 30.37 & 26.95   & 9.6M                     \\ \hline
WDVR~\cite{Fan2019empirical}                    & 23.47    & 27.36 & 25.84   & 30.11 & 26.69   & 1.2M                     \\ \hline
FRVSR~\cite{Sajjadi2018frame}                   & 23.02    & {\textcolor{red}{27.93}} & {\textcolor{red}{26.26}}   & 29.61 & 26.71   & 5.1M                     \\ \hline
RBPN~\cite{haris2019recurrent}                    & 23.96    & 27.74 & 26.21   & 30.70 & 27.15   & 12.7M                    \\ \hlineB{2}
GBR-WNN-S (Ours)        & 23.93    & 27.75 & 26.18   & 30.81 & 27.17   & 5.9M                     \\ \hline
GBR-WNN-M (Ours)        & {\textcolor{blue}{23.99}}    & 27.79 & 26.20   & {\textcolor{blue}{30.88}} & {\textcolor{blue}{27.21}}   & 8.8M                     \\ \hline
GBR-WNN-L (Ours)        & {\textcolor{red}{24.00}}    & {\textcolor{blue}{27.80}} & {\textcolor{blue}{26.22}}   & {\textcolor{red}{30.89}} & {\textcolor{red}{27.23}}   & 11.8M                    \\ \hline
\end{tabular}
\end{center}
\end{table*}

\begin{table*}[!t]
\caption{Quantitative comparison on Vid4 for 4${\times}$ video SR on RGB channel. {\textcolor{red}{Red}} and {\textcolor{blue}{blue}} indicates the best and the second best performance, respectively.\label{table2}}
\begin{center}
\begin{tabular}{|c||c|c|c|c||c||c|}
\hline
\multirow{2}{*}{Method} & \multicolumn{5}{c|}{PSNR (dB)}                    & \multirow{2}{*}{Params.} \\ \cline{2-6}
                        & Calendar & City  & Foliage & Walk  & Average &                          \\ \hline\hline
Bicubic                 & 18.96    & 23.75 & 22.21   & 24.94 & 22.47   & -                        \\ \hline
SOF-VSR~\cite{wang2020deep}                 & 20.96    & 25.43 & 24.01   & 27.80 & 24.55   & 1.0M                     \\ \hline
WAEN~\cite{choi2020wavelet}                    & 22.04    & 26.08 & 24.59   & 28.99 & 25.42   & 9.6M                     \\ \hline
WDVR~\cite{Fan2019empirical}                    & 21.75    & 25.84 & 24.44   & 28.74 & 25.19   & 1.2M                     \\ \hline
FRVSR~\cite{Sajjadi2018frame}                   & 21.37    & {\textcolor{red}{26.39}} & {\textcolor{red}{24.84}}   & 28.24 & 25.21   & 5.1M                     \\ \hline
RBPN~\cite{haris2019recurrent}                    & 22.17    & 26.21 & 24.78   & 29.31 & 25.62   & 12.7M                    \\ \hlineB{2}
GBR-WNN-S (Ours)        & 22.15    & 26.21 & 24.77   & 29.42 & 25.64   & 5.9M                     \\ \hline
GBR-WNN-M (Ours)        & {\textcolor{blue}{22.22}}    & 26.25 & 24.79   & {\textcolor{blue}{29.49}} & {\textcolor{blue}{25.69}}   & 8.8M                     \\ \hline
GBR-WNN-L (Ours)        & {\textcolor{red}{22.23}}    & {\textcolor{blue}{26.26}} & {\textcolor{blue}{24.81}}   & {\textcolor{red}{29.51}} & {\textcolor{red}{25.70}}   & 11.8M                    \\ \hline
\end{tabular}
\end{center}
\end{table*}
\begin{table*}[!t]
\caption{Analysis of adopted GBR framework and TWA module (Experiments here adopt a medium size model with 20 RBs) on Vid4 for 4${\times}$ video SR.
{\textcolor{red}{Red}} and {\textcolor{blue}{blue}} indicates the best and the second best performance, respectively.\label{table3}}
\begin{center}
\begin{tabular}{|c|c|c|c|c|c|}
\hline
\multicolumn{2}{|c|}{Model}                       & Test Model1 & Test Model2 & Test Model3         & GBR-WNN-M \\ \hline
\multicolumn{2}{|c|}{Params.}                     & 8.3M        & 8.3M        & 8.6M                & 8.8M      \\ \hline\hline
\multicolumn{2}{|c|}{Temporal Modeling Framework} & 2D CNN      & 2D CNN      & Uni-directional RNN & GBR       \\ \hline
\multicolumn{2}{|c|}{TWA Module}                  & {\xmark}    & {\cmark}    & {\cmark}            & {\cmark}  \\ \hlineB{2}
\multirow{2}{*}{PSNR (dB)}          & Y           & 26.62       & 26.80       & {\textcolor{blue}{26.92}}               & {\textcolor{red}{27.21}}     \\ \cline{2-6} 
                                    & RGB         & 25.11       & 25.28       & {\textcolor{blue}{25.40}}               & {\textcolor{red}{25.69}}     \\ \hline
\end{tabular}
\end{center}
\end{table*}

\begin{figure*}[!t]
\subfloat{\includegraphics[width=1.2in]{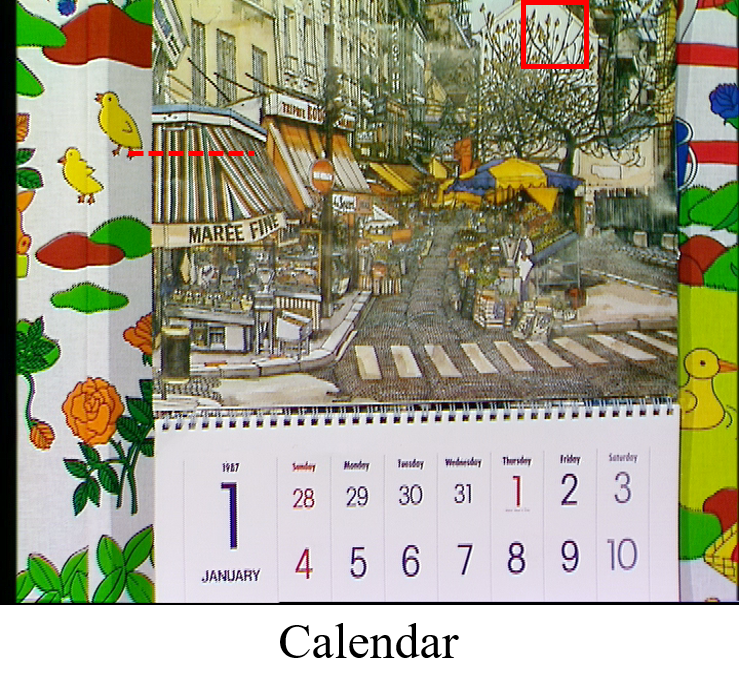}
\label{calendar_original}}
\hfil
\setcounter{subfigure}{0}
\subfloat[Bicubic.]{\includegraphics[width=0.9in]{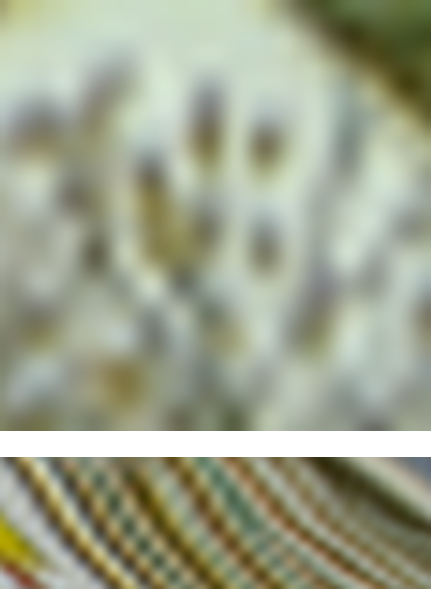}
\label{calendar_bicubic}}
\hfil
\subfloat[SOF-VSR~\cite{wang2020deep}.]{\includegraphics[width=0.9in]{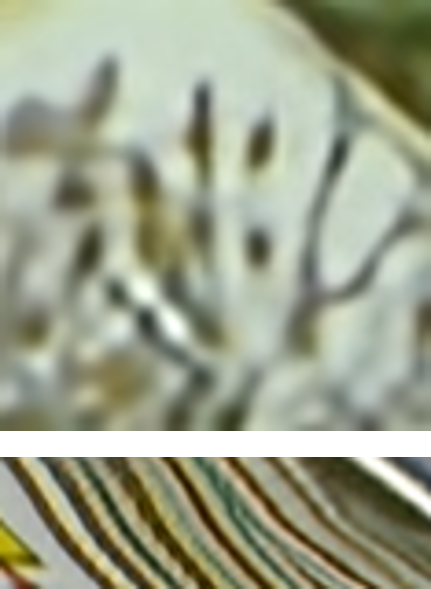}
\label{calendar_sofvsr}}
\hfil
\subfloat[WAEN~\cite{choi2020wavelet}.]{\includegraphics[width=0.9in]{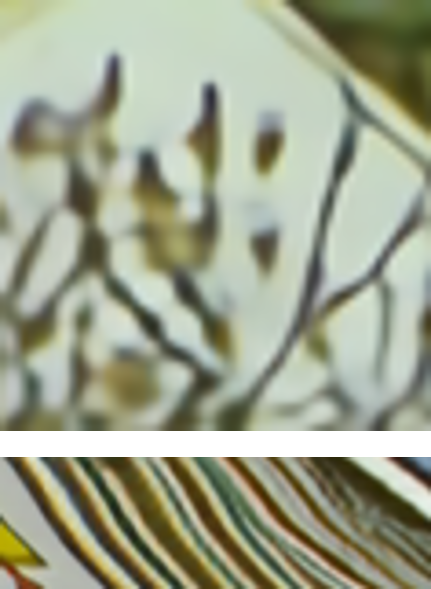}
\label{calendar_waen}}
\hfil
\subfloat[WDVR~\cite{Fan2019empirical}.]{\includegraphics[width=0.9in]{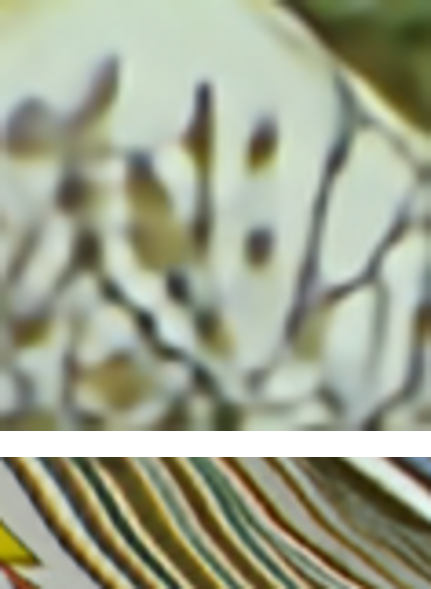}
\label{calendar_wdvr}}
\\[-2ex]
\hspace*{.22\textwidth}\quad
\subfloat[FRVSR~\cite{Sajjadi2018frame}.]{\includegraphics[width=0.9in]{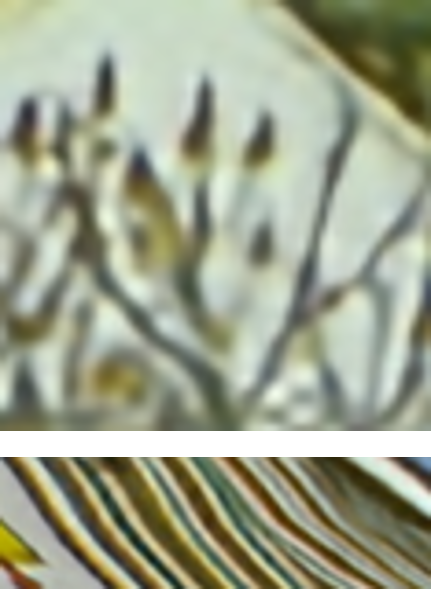}
\label{calendar_frvsr}}
\hfil
\subfloat[RBPN~\cite{haris2019recurrent}.]{\includegraphics[width=0.9in]{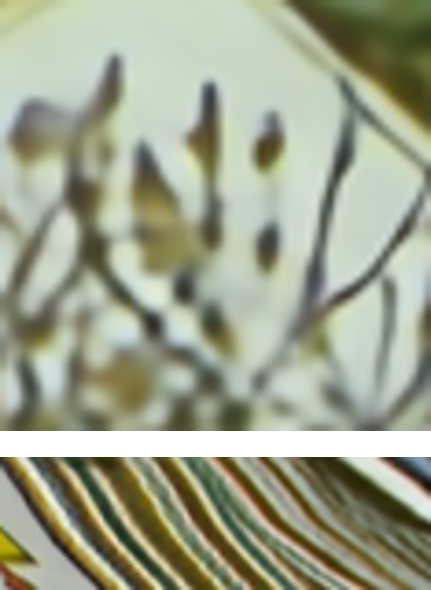}
\label{calendar_rbpn}}
\hfil
% \subfloat[GBR-WNN-S.]{\includegraphics[width=0.8in]{calendar_gbr-wnn-s}
% \label{calendar_gbr-wnn-s}}
% \hfil
% \subfloat[GBR-WNN-M.]{\includegraphics[width=0.8in]{calendar_gbr-wnn-m}
% \label{calendar_gbr-wnn-m}}
% \hfil
\subfloat[GBR-WNN-L.]{\includegraphics[width=0.9in]{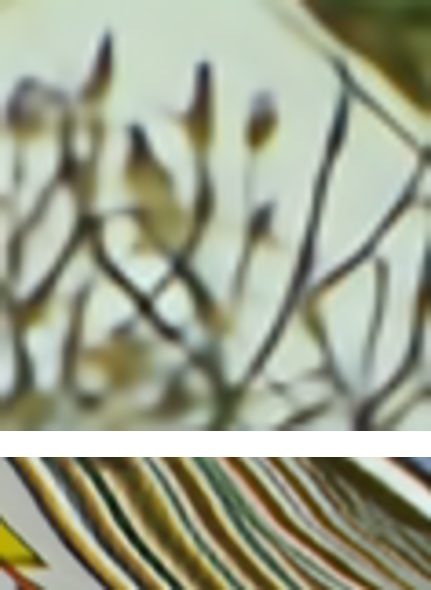}
\label{calendar_gbr-wnn-l}}
\hfil
\subfloat[GT.]{\includegraphics[width=0.9in]{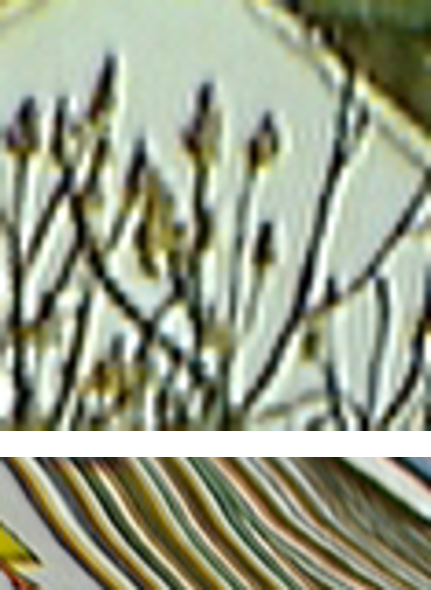}
\label{calendar_gt}}
\\[-2ex]
\subfloat{\includegraphics[width=1.2in]{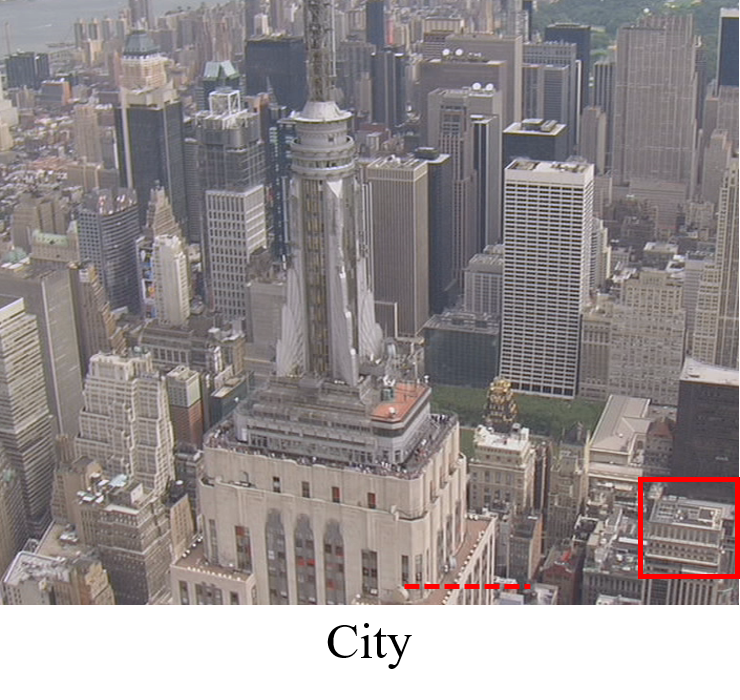}
\label{city_original}}
\hfil
\setcounter{subfigure}{0}
\subfloat[Bicubic.]{\includegraphics[width=0.9in]{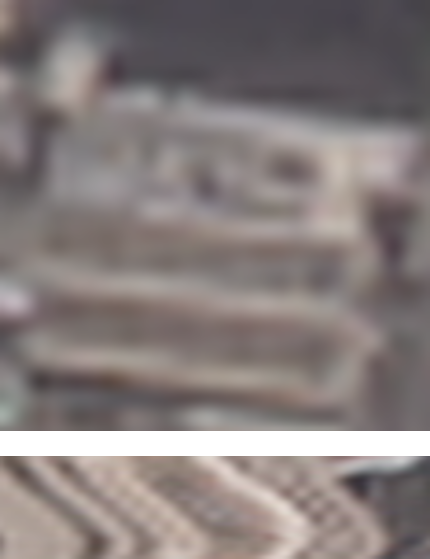}
\label{city_bicubic}}
\hfil
\subfloat[SOF-VSR~\cite{wang2020deep}.]{\includegraphics[width=0.9in]{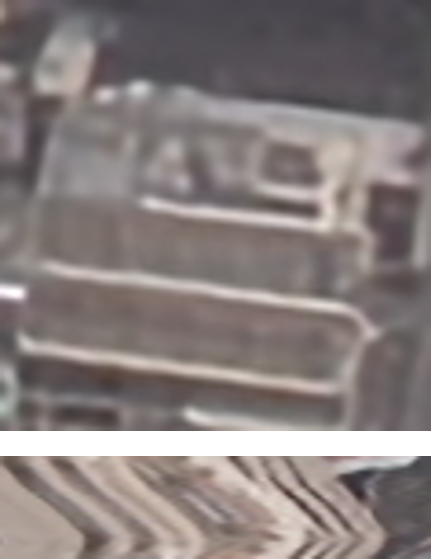}
\label{city_sofvsr}}
\hfil
\subfloat[WAEN~\cite{choi2020wavelet}.]{\includegraphics[width=0.9in]{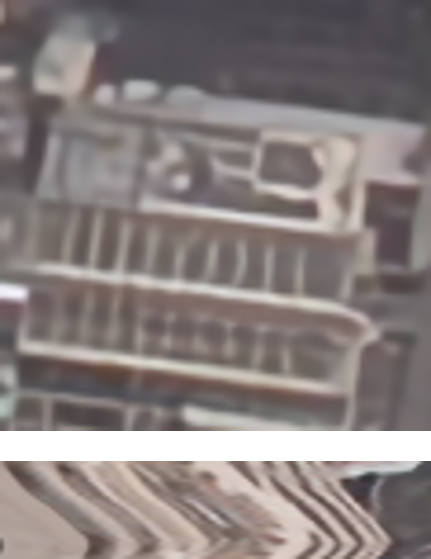}
\label{city_waen}}
\hfil
\subfloat[WDVR~\cite{Fan2019empirical}.]{\includegraphics[width=0.9in]{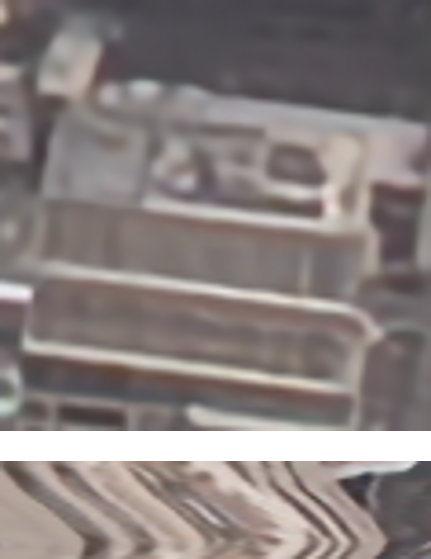}
\label{city_wdvr}}
\\[-2ex]
\hspace*{.22\textwidth}\quad
\subfloat[FRVSR~\cite{Sajjadi2018frame}.]{\includegraphics[width=0.9in]{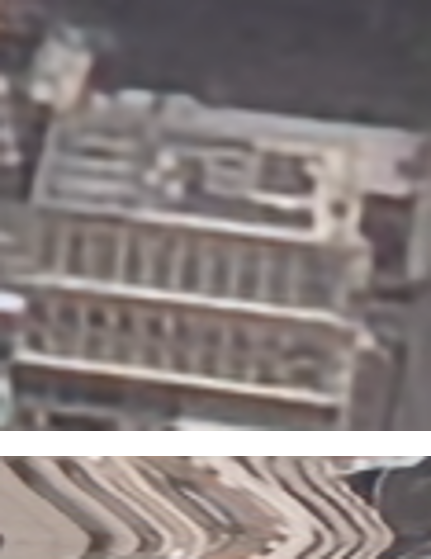}
\label{city_frvsr}}
\hfil
\subfloat[RBPN~\cite{haris2019recurrent}.]{\includegraphics[width=0.9in]{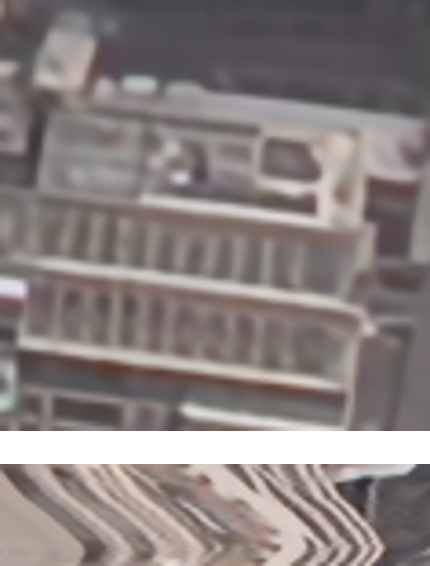}
\label{city_rbpn}}
\hfil
% \subfloat[GBR-WNN-S.]{\includegraphics[width=0.8in]{city_gbr-wnn-s}
% \label{city_gbr-wnn-s}}
% \hfil
% \subfloat[GBR-WNN-M.]{\includegraphics[width=0.8in]{city_gbr-wnn-m}
% \label{city_gbr-wnn-m}}
% \hfil
\subfloat[GBR-WNN-L.]{\includegraphics[width=0.9in]{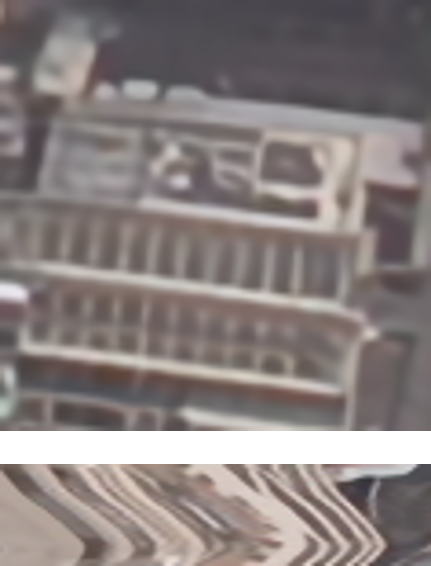}
\label{city_gbr-wnn-l}}
\hfil
\subfloat[GT.]{\includegraphics[width=0.9in]{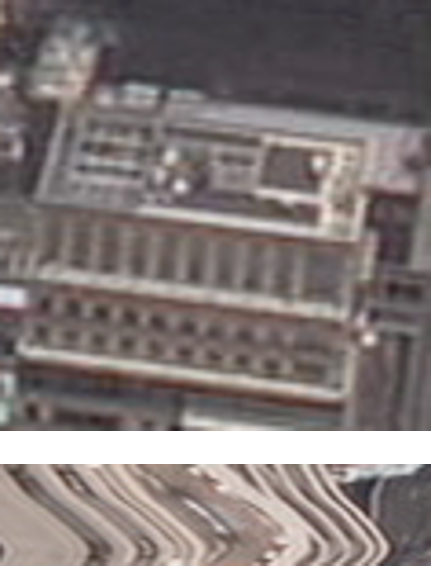}
\label{city_gt}}
\caption{Qualitative comparison on Vid4 for 4${\times}$ video SR. Zoom in to see better visualization.}
\label{fig5}
\end{figure*}

%-------------------------------------------------------------------------
\subsection{Comparison with the-state-of-the-arts}
We compare the proposed GBR-WNN with Bicubic and several state-of-the-art methods including all kinds of temporal modeling frameworks, (1) 2D CNN: the optical flows for VSR (SOF-VSR)~\cite{wang2020deep} and the wavelet attention embedding network (WAEN)~\cite{choi2020wavelet}, (2) 3D CNN: the wide-activated 3D convolutional network for video restoration (WDVR)~\cite{Fan2019empirical}, and (3) RNN: the recurrent back-projection network (RBPN)~\cite{haris2019recurrent} and the frame-recurrent VSR (FRVSR)~\cite{Sajjadi2018frame} on Vid4 dataset.

For SOF-VSR~\cite{wang2020deep} and RBPN~\cite{haris2019recurrent}, we used a provided pre-trained model to produce their results. For WAEN~\cite{choi2020wavelet} and WDVR~\cite{Fan2019empirical}, we trained the model using Vimeo-90K dataset.
We produced the result of FRVSR~\cite{Sajjadi2018frame} using the super-resolved Vid4 result images provided by the authors because the source code and the pre-trained model were not opened.
In this paper, the first two frames and the last two frames are not used for overall performance evaluation because the FRVSR~\cite{Sajjadi2018frame} provides excluding these four frames.

The quantitative results in term of PSNR on Y (luminance) and RGB channels are shown in Table~\ref{table1} and Table~\ref{table2}, respectively.
We also show the number of parameters for each model.
In our experiments, we evaluated our GBR-WNN with different number of residual blocks (RBs) in the reconstruction module.
The large size model consisting of 30 RBs (GBR-WNN-L), medium size model with 20 RBs (GBR-WNN-M) and small size model with 10 RBs (GBR-WNN-S) were tested.
The results show the tendency that the model with increased RBs performs better.

Comparing with other methods, our GBR-WNN-L shows the best performance and GBR-WNN-M is in the second place on average in both Y and RGB channels.
% When comparing our GBR-WNN-L with RBPN~\cite{haris2019recurrent} which has the highest performance among the existing methods, we can see that the proposed GBR-WNN-L can brings about 0.08dB higher PSNR performances on average in both Y and RGB channels.
Even though RBPN~\cite{haris2019recurrent} needs 0.9M additional parameters, our GBR-WNN-L outperforms RBPN~\cite{haris2019recurrent} in term of PSNR.
Furthermore, when comparing our small size model GBR-WNN-S with RBPN, our method has better average performances.

Although FRVSR~\cite{Sajjadi2018frame} has better performances on City and Foliage clips in both Y and RGB channels, the gap between our GBR-WNN-L and FRVSR~\cite{Sajjadi2018frame} is small in this clips.
On the contrary, the proposed GBR-WNN-L outperforms FRVSR by a large margin on Calendar and Walk clips.
Comparing with WAEN using DWT-based feature extractor, our methods achieves better performance.
It means that the proposed GBR framework is more effective in estimating the sophisticated temporal feature than general 2D CNN temporal modeling framework.

The qualitative results on Vid4 dataset for Calendar and City clips are presented in Figure~\ref{fig5}.
For result of each method, the upper row means zoomed a visual area and the lower row means temporal profile.
For the temporal profiles, we used 144th line for Calendar clip and 550th line for City clip in the entire frame of the sequences.
Our GBR-WNN-L recovers more accurate textures with more smooth temporal transition compared to the existing methods.

For a more detailed analysis of the proposed GBR-WNN, Table~\ref{table3} shows the results as adopted our GBR framework and TWA module in our method.
From this result, we can explain that combination of two proposed core components produces better performance than using a single component.
By comparing Test Model2 to Test Model1, we can verify that the TWA module exploits useful information in VSR problem.
When comparing GBR-WNN-M with Test Model2 and Test Model3, we can see that the GBR framework is very efficient and beneficial in dealing with the sequential data.

%-------------------------------------------------------------------------

%-------------------------------------------------------------------------
\section{Conclusion}
In this paper, we have proposed a group-based bi-directional recurrent wavelet neural networks (GBR-WNN) for VSR.
The proposed method consists of two core components: group-based bi-directional RNN (GBR) temporal modeling framework and temporal wavelet attention (TWA) module.
The temporal continuities between multiple frames were aligned with the management structure based on the group of pictures (GOP) in the proposed GBR framework.
With the enhanced temporal modeling framework, the proposed network can recover precise temporal details improving SR performance.
In addition, the proposed TWA module was able to make full use of spatio-temporal information among consecutive frames to generate the enriched features.
We have compared the performance of proposed method with other recent state-of-the-art VSR approaches.
Experimental results demonstrated that the proposed GBR-WNN could obtain better quality in VSR.

%-------------------------------------------------------------------------

{\small
\bibliographystyle{ieee_fullname}
\bibliography{egbib}
}

\end{document}